\documentclass[sigconf]{acmart}
\usepackage[T1]{fontenc}
\usepackage{array}
\usepackage{amsmath}
\usepackage{graphicx}
\usepackage{hyperref}
\usepackage[utf8]{inputenc}
\AtBeginDocument{%
  \providecommand\BibTeX{{%
    \normalfont B\kern-0.5em{\scshape i\kern-0.25em b}\kern-0.8em\TeX}}}

\begin{document}

\title{Does Adversarial Oversampling Help us?}

\author{Tanmoy Dam}
\affiliation{%
  \institution{School of Engineering and Information Technology, University of New South Wales Canberra}
  \city{Canberra}
  \country{Australia}
  \postcode{2612}
}
\email{t.dam@student.adfa.edu.au}

\author{Md Meftahul Ferdaus}
\affiliation{%
  \institution{Institute for Infocomm Research, Agency for Science, Technology and Research (A*STAR)}
  \country{Singapore}
  \postcode{138632}}
\email{ferdaus\_meftahul@i2r.a-star.edu.sg}

\author{Sreenatha G. Anavatti}
\affiliation{%
  \institution{School of Engineering and Information Technology, University of New South Wales Canberra}
  \city{Canberra}
  \country{Australia}
  \postcode{2612}}
\email{agsrenat@adfa.edu.au}

\author{Senthilnath Jayavelu}
\affiliation{%
  \institution{Institute for Infocomm Research, Agency for Science, Technology and Research (A*STAR)}
  \country{Singapore}
  \postcode{138632}}
\email{J_Senthilnath@i2r.a-star.edu.sg}

\author{Hussein A. Abbass}
\affiliation{%
  \institution{School of Engineering and Information Technology, University of New South Wales Canberra}
  \city{Canberra}
  \country{Australia}
  \postcode{2612}}
\email{hussein.abbass@gmail.com}

  
  
  




\begin{abstract}

Traditional oversampling methods are generally employed to handle class imbalance in datasets. This oversampling approach is independent of the classifier; thus, it does not offer an end-to-end solution. To overcome this, we propose a three-player adversarial game-based end-to-end method, where a domain-constraints mixture of generators, a discriminator, and a multi-class classifier are used. Rather than adversarial minority oversampling, we propose an adversarial oversampling (AO) and a data-space oversampling (DO) approach. In AO, the generator updates by fooling both the classifier and discriminator, however, in DO, it updates by favoring the classifier and fooling the discriminator. While updating the classifier, it considers both the real and synthetically generated samples in AO. But, in DO, it favors the real samples and fools the subset class-specific generated samples. To mitigate the biases of a classifier towards the majority class, minority samples are over-sampled at a fractional rate. Such implementation is shown to provide more robust classification boundaries. The effectiveness of our proposed method has been validated with high-dimensional, highly imbalanced and large-scale multi-class tabular datasets. The results  as measured by average class specific accuracy (ACSA) clearly indicate that the proposed method provides better classification accuracy (improvement in the range of 0.7\% to 49.27\%) as compared to the baseline classifier.

\end{abstract}



\maketitle

\section{Introduction}
Traditional machine learning (ML) algorithms perform well in comparatively balanced datasets. However, in real-world problems such as fraud detection, fault detection, medical diagnosis, and dense object detection~\cite{mullick2019generative}, all the classes do not usually have an equal number of samples. In such cases, ML algorithms try to search for the best decision boundaries. However, minority classes may face challenges resultant from their low frequencies, which could lead to misleading decision boundaries. Unless a cost-sensitive classification approach is followed, ML approaches do not handle a skewed data distribution well and bias their choices to the overall accuracy of the classifier, which in turn favours majority classes. Moreover, a further challenge exists when minority samples are drawn from a biased subset of the class distribution, which causes unrepresentative sampling. The complexity increases when dealing with multi-class imbalanced classification problems. A class can be considered as a majority with respect to some other class, while a minority or well-balanced one in contrast with the remaining classes. 

Methods devised by the ML community to deal with class imbalance problem can be divided into two broad types, namely data-level and algorithm-level methods. In the latter methods, cost-sensitive learning is utilized where the classifiers are penalized more heavily for miss-classifying the minority class more than the majority class. In data-level methods (oversampling and undersampling \cite{he2009learning}), data may be re-sampled randomly from the original data, or maybe sampled synthetically by generating synthetic samples. These methods can be applied in the data pre-processing stage, which provides them more flexibility than algorithm-level methods \cite{japkowicz2000class}. 

In recent times, generative adversarial networks (GANs) \cite{goodfellow2014generative} have been widely used in handling imbalance problems due to their data generation capabilities \cite{douzas2018effective, mullick2019generative}. In GANs, two networks, namely generator and discriminator, contest with each other in an adversarial game. The generator takes lower-dimensional latent space to generate a realistic complex data distribution. The discriminator discriminates between real and generated samples. The efficacy of generated synthetic samples to handle imbalance problems in the tabular dataset was introduced by class-conditional GANs method \cite{douzas2018effective}. Due to majority class dominance, class-conditional GANs are known to suffer from mode collapse problems. Therefore, generative adversarial minority oversampling (GAMO) \cite{mullick2019generative} was introduced to tackle mode collapse by considering domain constraints convex generators. The generators generate the minority class near decision boundaries, which are difficult for classifiers to classify.  

Usually, cross-entropy (CE) loss-based networks perform better than the least square (LS)-based ones in classification tasks \cite{englesson2021generalized}. However, the GAMO classifier is based on the least square (LS) method, where the CE based approach does not improve enough the classification performance than the baseline due to the generators were not updated through complement cross-entropy loss (CCE) \cite{chen2019complement} in GAMO. Therefore, we propose two adversarial oversampling methods that cover all the possible combinations of the game between classifier, generator and discriminator. The contributions of this paper are as follows: 
\begin{itemize}
  \item Our proposed method is based on a three-player network: domain constraints class conditionals mixture of generators ($G$), a discriminator ($D$) and a classifier ($Q$). Unlike the adversarial minority oversampling method, utilized in benchmark three players network called GAMO, we propose an adversarial oversampling (AO) and a data-space oversampling (DO) approach.      
  \item In AO, the $G$ is updated by fooling $Q$ and $D$ simultaneously. That means, $G$ will enforce to generate the samples within class distribution to fool the classifiers ($Q$). Both $Q$ and $D$, simultaneously enforce generators to generate more prominent class samples near decision boundaries. $Q$ will be updated through oversampling of class distributions to mitigate majority class distributions. 
  \item In DO, the $G$ is updated by fooling only $D$ but in favour of $Q$, whereas, $Q$ is updated by favouring of real samples and fooling of subset class-specific generated samples. That means, class specific subset samples are generated by $G$ to give more robust classification boundaries. In ablations studies, we observed that each fractions of data space oversampling can improve the classification performance.   
\end{itemize}

\section{Problem formulation}
Let us assume the joint distribution of training set $(X, Y)$ of which $N_t$ training samples are independent and identically distributed (i.i.d) over $C$  classes. The $k$-th training sample, $(\bf {x}_k, \bf{y}_k)$ represents the attribute information $(\bf{x}_k \in \mathbb{R}^d)$ and class information $(\bf{y}_k \in \mathbb{R}^C)$. The total number of samples can be presented as $N = \sum_{l=c}^{C} p_l$, where, $p_l$ is the size of $l$-th class. In real situations, distributed class ($p_k$) is not uniform. In general, the imbalance class variances are represented as $\{p_1,p_2,...,p_c,...,p_C \}$ where $p_C$ is the largest class, which can be 50 times bigger than the smallest class i.e. $p_C \geq 50*p_1$. The objective is to design a $(Q)$ that can estimate the $Q(x^*_k)= \hat y \in y_k$ but designing such $Q(.)$ to estimate the underlying distribution of training data under imbalance settings is difficult.   

\section{Adversarial Over-Sampling}
We have proposed generative model-based two oversampling approaches, namely, adversarial oversampling (AO) and data space oversampling (DO), where samples generated by G support to improve the performance of Q. In AO, our domain constraints class-conditional generators $G$ will generate samples at different localities of the data-space by obtaining gradients from $D$ and $Q$ by fooling both $D$ and $Q$ simultaneously. The gradients of $Q$ are updated through real data and adversary over-sampled data to balance the decision boundaries of $Q$. Like the basic GAN \cite{goodfellow2014generative}, $D$ is updated to maximize the probability of assigning the real data and samples generated from $G$. In DO, the gradients of $G$ are updated by fooling $D$ and favouring $Q$ simultaneously. However, $Q$ is updated through the real samples and generated class conditionals samples to fool $Q$. In DO, the $D$ is updated similarly as in AO. To mitigate the imbalance impact at $Q$, data-space, over-sampled at a different fractional rate, is defined to observe the impact of classification performance.  

\subsection{Mixture of generators}
\begin{figure}
    \centering
    \includegraphics[scale=0.5]{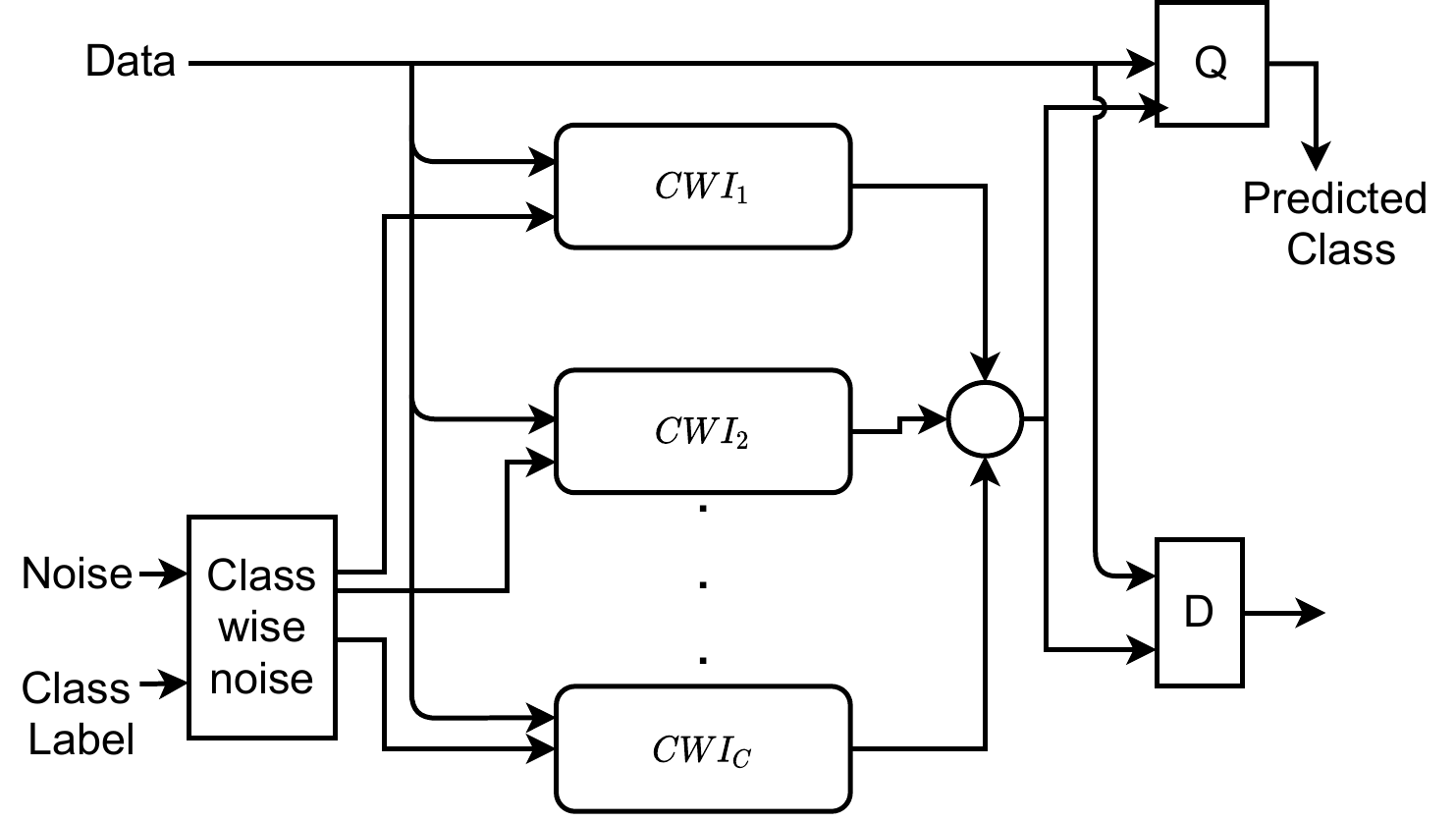}
    \caption{The mixture of generators takes class conditionals real data($\tau_{k} \in X $) as well as class-embedded($C$) noise prior($z$).  CWI block selects class specific real instance through Q and D. Selection criteria is based on two strategies: AO and DO}
    \vspace{-2mm}
    \label{fig:my_archi}
\end{figure}

When no constraint is considered at individual class generators' outputs, the major class intervenes into minor classes domain even if the noise samples are taken from different Gaussian distributions \cite{mullick2019generative,douzas2018effective}. In GAMO, class-specific real data is considered at the generators, which confirms the generators' ability to generate minor classes at minor-class distributions. Instead of generating real like data, the class conditional generators are used to generate latent vectors which select the instances (samples) of a specific class of real distributions. This ensured that all generated samples remained within minor-class distributions. Similar to GAMO, our generator is also designed to generate samples to retain all the class distributions. Both methods are based on adversarial over-samplings, therefore, we need to design a mixture of all class-conditional generators. Our mixture of generators consist of one module: class-specific weight instance (CWI) as shown in Fig. \ref{fig:my_archi}, which takes Gaussian normal distribution latent variable $(z)$ to map $d$-dimensional real value $(0,1)$. It is obtained by using softmax activation of the last layer. The output of CWI unit is $g(z|k)\in \mathbb{R}^d$ since $g(z|k) \geq 0$ and $(g(z|k))^T*\bf{1} =1$. Thus, CWI will select an instance of $X_k$ class-data. The $k$-th class conditional generator's output is obtained by considering the average weight of $k$-th real data distribution, which can be expressed as: 
\begin{equation}
    G(z|k)=(g(z|k))^T\tau_k , \quad \tau_k \subset X
\end{equation}
where, $\tau_k$ is denoting all the samples of $k$-th class belonging to $X$. 

Finally, the adversarial game playing between three players leads to the following optimisation problem:
\begin{equation}
    \min_{G, Q}\max_{D} L_{\{AO/DO \}}(G,D,Q) 
\end{equation}
where, $L_{AO}= L_{AO}^G + L_{AO}^D + L_{AO}^Q$ \\
and  $ L_{DO}= L_{DO}^G + L_{DO}^D + L_{DO}^Q $ 

where,
\begin{multline}
    L_{AO}^G = \mathbb{E}_{G(z|i) \sim p_{i}^g} (s(1-D(G(z|i)))) + \\ \mathbb{E}_{G(z|i) \sim p_{i}^g} (log(1-Q(G(z|i)))) \\
             = ... + CCE(p_i,Q(G(z|i)))
\end{multline}

\begin{align}
    L_{AO}^D &= \mathbb{E}_{x \sim p_{k}^r} (s(D(x)) + \mathbb{E}_{G(z|k) \sim p_{k}^g} (s(1-D(G(z|k))))\\
    L_{AO}^Q &=  \mathbb{E}_{x \sim p_{k}^r} (log(Q(x)) + \mathbb{E}_{G(z|j) \sim p_{j}^g} (log(Q(G(z|j))))
\end{align}

Similarly,
\begin{align}
    L_{DO}^G  = & \mathbb{E}_{G(z|i) \sim p_{i}^g} (s(1-D(G(z|i)))) + \nonumber \\ & \mathbb{E}_{G(z|i) \sim p_{i}^g} (log(Q(G(z|i))))\\
    L_{DO}^D = &\mathbb{E}_{x \sim p_{k}^r} (s(D(x)) + \mathbb{E}_{G(z|k) \sim p_{k}^g} (s(1-D(G(z|k))))\\
    L_{DO}^Q  =  &\mathbb{E}_{x \sim p_{k}^r} (log(Q(x)) +  \mathbb{E}_{G(z|k) \sim p_{k}^g} (log(1-Q(G(z|k)))) \nonumber\\
              =& ... + CCE(p_k,Q(G(z|k))) 
\end{align}

where, $log$ term is associated with CE loss. The $p_{k}^r$ and $p_{k}^g$ belong to the real and generated distribution for $k$-th class. In addition, the $p_{i}^g$ and $p_{j}^g$ belong to the generated samples that drawn from balanced distribution and inverse of real distributions respectively.

The operator function $s(.)$  selects different GANs. The operator $(s)$ is defined as  $s(x)=logx$ for vanilla GANs \cite{goodfellow2014generative} and $s(x)=x$ for Wasserstein GANs (WGANs) \cite{gulrajani2017improved}. Due to the stability nature of GANs game, we have used $1$-gradient penalty (gp) based WGANs for validating both the methods.
 
\vspace{-2mm}

\section{Experimental Validations}
Six different numerical multi-class datasets are taken from UCI machine learning repository \cite{lichman2013uci} and KAGGLE competition \footnote{https://www.kaggle.com/c/GiveMeSomeCredit} for validating the proposed methods. The detailed description of the datasets is given in Table\ref{Table:dataset_details}. The multi-class datasets are chosen based on three main key factors, such as high-dimensions (Secom and Wafer), high imbalance ratio (IR) (Secom, Credit and Yeast), and large-scale (credit-card). For handling the missing data, we have used the KNN imputation technique. Secom is the only case study that has missing values in each attribute, which are filled up by setting $2$-nearest neighboring values. We measure the performance of the baseline classifier ($Q$) by two indices: Average Class Specific Accuracy (ACSA) and Geometric Mean (GM) \cite{mullick2019generative}. 

To make the experimental studies more reliable, we divided the datasets randomly so that $70 \%$ were used for training and $30 \%$ were used for testing. Thus, for the Secom dataset, the training  and testing samples are  $\{1023, 73\}$ and $\{440, 31\}$ respectively. Similarly, for Prima dataset, class distributions are $\{401, 213\}$ and $\{99, 55\}$ respectively. In our case studies, in all the datasets except Wafer, the minor class present in training samples are more than the minor class present in testing samples. However, in real-life scenarios, the testing samples may be more than training samples.  Thus, the Wafer dataset is a good candidate which falls under this category with the training  and testing samples being $\{97,  903 \}$ and  $\{665,  5499\}$ respectively. Besides two-class classification problems, we have chosen a ten-class dataset (Yeast dataset) for checking the superiority of the proposed method. In the Yeast dataset, the train-test ratio is $(50-50 \%)$, where the data are shuffled randomly. 

\begin{table}
\caption{Detail of datasets ($R^d:$ Data-dimensions, $IR$: Imbalance ratio, $C$: number of classes)}

\begin{centering}
\begin{tabular}{>{\centering}p{0.99cm}>{\centering}p{0.8cm}c>{\centering}p{2.6cm}>{\centering}p{0.6cm}}
\hline 
Datasets & $R^d$ & \# Samples ($N$) & $IR$ & $C$\tabularnewline
\hline 
Secom & 590 & 1567 & 1463:104 (14.07) & 2\tabularnewline
Prima & 8 & 768 & 500:268 (1.87) & 2\tabularnewline
Haberman & 3 & 306 & 225:81 (2.78) & 2\tabularnewline
Wafer & 152 & 7164 & 6402:762 (8.40) & 2\tabularnewline
Credit & 10 & 150000 & 139974:10026 (13.96) & 2\tabularnewline
Yeast & 8 & 1484 & 463:5 (92.6) & 10\tabularnewline
\hline 
\end{tabular}
\par\end{centering}
\label{Table:dataset_details}
\end{table}


\begin{table*}[]
\caption{Classification performance on Experimental datasets}
\scalebox{0.8}{
\begin{tabular}{lllllllllllllllllll}
Datasets & \multicolumn{2}{l}{Q} & \multicolumn{2}{l}{RO+Q} & \multicolumn{2}{l}{COT} & \multicolumn{2}{l}{SMOTE+Q} & \multicolumn{2}{l}{B-SMOTE+Q} & \multicolumn{2}{l}{ADASYN+Q} & \multicolumn{2}{l}{GAMO} & \multicolumn{2}{l}{AO} & \multicolumn{2}{l}{DO} \\ \hline
         & ACSA      & GM        & ACSA        & GM         & ACSA       & GM         & ACSA         & GM           & ACSA          & GM            & ACSA          & GM           & ACSA        & GM         & ACSA       & GM        & ACSA       & GM        \\ \hline
Secom    & 57.06     & 42.82     & 58.99       & 43.69      & 60.51      & 49.57      & 55.57        & 42.15        & 55.08         & 35.42         & 59.12         & 46.47        & 61.17       & 56.90      & 65.82      & 65.36     & \bf{68.84}      & \bf{65.73}     \\
Prima    & 76.89     & 76.49     & 73.09       & 73.09      & 76.55      & 76.53      & 74.10        & 71.94        & 76.82         & 75.34         & 76.20         & 75.16        & 70.10       & 68.78      & 77.42      & 77.00     & \bf{77.46}      & \bf{77.45}     \\
Haberman & 65.00     & 60.91     & 70.00       & 69.48      & 68.84      & 67.93      & 69.61        & 69.48        & \bf{70.38}         & \bf{70.20}         & 69.61         & 69.48        & 58.62       & 50.46      & 60.00      & 58.38     & \bf{70.38}      & \bf{70.20}     \\
Wafer    & 98.82     & 98.81     & 97.82       & 97.81      & 97.81      & 97.80      & 95.30        & 95.22        & 88.07         & 87.27         & 95.65         & 95.57        & 98.78       & 98.77      & 98.99      & 98.99     & \bf{99.44}      & \bf{99.44}     \\
Credit   & 50.61     & 11.51     & 50.04       & 3.08       & 51.84      & 19.83      & 50.00        & 00.48        & 50.00         & 00.00         & 50.00         & 0.84         & 61.46       & 61.36      & 61.65      & 62.19     & \bf{75.55}     &  \bf{75.54}     \\ \hline
Yeast    & 61.67     & 00.00     & 60.20       & 00.00      & 57.17      & 00.00      & 57.27        & 45.52        & 55.85         & 43.37         & NA            & NA           & 57.52       & 00.00      & 51.29      & 00.00     & \bf{62.81}      & \bf{53.77}   \\ \hline 
\multicolumn{6}{c}{\footnotesize Note: best outcome for each metric in \textbf{bold}. NA-Not Applicable}
\end{tabular}}
\label{Table:experimental_results}
\end{table*}
\vspace{-2mm}
\subsection{Existing methods}
We have compared the proposed oversampling approaches with two types of oversampling methods that exist in the literature, such as the two-stages method and the one-stage method. Some popular two-stages methods are SMOTE \cite{chawla2002smote}, Borderline-SMOTE (B-SMOTE) \cite{han2005borderline}, and ADSYN \cite{he2008adasyn} where these methods are applied to balance the dataset. Same configurations for different parameters as recommended  by each corresponding author \cite{chawla2002smote} \cite{han2005borderline}  \cite{he2008adasyn} are maintained here. Thus, we have considered the five nearest neighboring points to create the synthetic samples except for the Yeast dataset, in which the minor classes are too low to follow the baseline recommendations. Hence, we have taken the minimum nearest neighboring points which support the above methods. Recently developed one stage method like GAMO \cite{mullick2019generative} is also considered for comparison. We have also compared with the complementary objective training (COT) \cite{chen2019complement} to validate the effectiveness of the proposed methods.
\subsection{Simulation results and Discussions}
Table\ref{Table:experimental_results} describes the classification performance for the different datasets using the ACSA and GM values \cite{mullick2019generative}. Here, ACSA \cite{huang2016learning,wang2017learning,mullick2019generative} can be defined as $ACSA=0.5(t_p/N_p+t_n/N_n)$, where $N_p$ and $N_n$ are the numbers of positive and negative samples, while $t_p$ and $t_n$ are the numbers of true positive and true negative. For fairer comparison with existing methods, we have used a testing set after each epoch of training. A similar approach has also been followed for GAMO. Since our experimental studies are based on tabular datasets, we have used $3$-layered multi-layer perceptron (MLP) models for $D,G$ and $4$-layered MLP model for $Q$. All the methods are trained on $100$ epochs. 

\begin{figure}[h]
\includegraphics[scale=0.4]{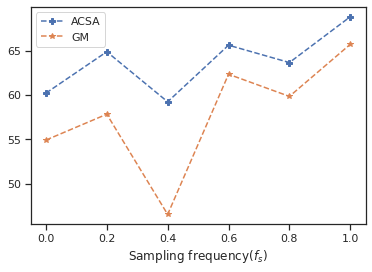}
\caption{Performance of $Q$ at different sampling frequency rate}
\vspace{-2mm}
\label{fig:DO_oversampling_frequency}
\end{figure}

The simulation results listed in Table\ref{Table:experimental_results} indicates that our DO approach is the best suitable candidate while dealing with all the key factors. For better understanding of DO at different over-sampling frequency rate ($f_s = \frac{p_1}{p_C}$), we have illustrated the performance on Secom dataset in Fig. \ref{fig:DO_oversampling_frequency}. Besides DO, our other proposed approach i.e. AO out-performs the baseline $Q$ while considering high-dimensional and highly-imbalance datasets like Secom, Wafer and Credit dataset. The AO also outperformed GAMO in almost all cases except for the Yeast dataset. This shows that AO with CE loss is a better option than the GAMO's minority oversampling with LS loss. When we  considered classical random oversampling technique $(RO+Q)$ (that is independent of the baseline classifier), it does not guarantee performance improvement in the baseline classifier. Rather, it declines baseline performance in some cases as over-fitting the model parameters, which occurs due to the generation of samples in regions which may not be helpful to $Q$. Thus, SMOTE, B-SMOTE and ADSYN, are not good options all the time. It is also observed from Table \ref{Table:experimental_results} that COT is not a suitable option for handling imbalanced classification problem. Thus, for handling highly imbalanced and high-dimensional big datasets, the proposed DO-based single-stage method is performing better than not only the two-stage classical oversampling methods but also the recently developed single-stage variants.   
\vspace{-2mm}

\section{Conclusion and Future Work}
This paper is based on the adversarial game between G, D and Q to handle imbalanced datasets. From the experimental results on various numerical datasets, it is observed that the proposed DO approach outperforms AO as well as alternative methods. The proposed DO methods can be applied on high-dimensional, high-IR, large scale datasets spanning over binary-class to multi-class imbalance problems. For better understanding the effectiveness of the proposed DO method, we will be experimenting on image dataset in the near future, where deep neural network will be applied.

\begin{acks}
Tanmoy Dam acknowledges the scholarship from the University of New South Wales namely University International Postgraduate Award (UIPA).
Md Meftahul Ferdaus and Senthilnath Jayavelu acknowledge funding from the Accelerated Materials Development for Manufacturing Program at A*STAR via the AME Programmatic Fund by the Agency for Science, Technology and Research under Grant No. A1898b0043.
\end{acks}

\bibliographystyle{ACM-Reference-Format}
\bibliography{sample-base}

\end{document}